\documentclass[conference]{IEEEtran}
\usepackage{times}
\usepackage{times}

\usepackage[numbers]{natbib}
\usepackage{multicol}
\usepackage[bookmarks=true]{hyperref}
\usepackage{xspace}
\usepackage{amsmath}
\usepackage{amssymb}

\usepackage{siunitx}
\usepackage{graphicx}
\usepackage{subcaption}
\usepackage[numbers]{natbib}
\usepackage{booktabs}       

\DeclareMathOperator*{\argmax}{argmax}

\usepackage{comment}

\usepackage{algorithm} 
\usepackage{algorithmic}  
\usepackage[algo2e]{algorithm2e} 
\usepackage{bm}

\usepackage{xcolor}

\let\labelindent\relax
\usepackage{enumitem}

\usepackage{algorithm2e}

\newcommand{\xxnote}[3]{}
\ifx\hidenotes\undefined
  \usepackage{color}
  \renewcommand{\xxnote}[3]{\color{#2}{#1: #3}}
\fi
\usepackage{url}

\begin{document}


\title{A Quality Diversity Approach to Automatically \\ Generating Human-Robot Interaction Scenarios \\ in Shared Autonomy}

\author{
\authorblockN{
Matthew C. Fontaine and 
Stefanos Nikolaidis}
\authorblockA{Department of Computer Science\\
University of Southern California\\
Los Angeles, CA, USA\\
mfontain@usc.edu, nikolaid@usc.edu}
}

\maketitle

\begin{abstract}
The growth of scale and complexity of interactions between humans and robots highlights the need for new computational methods to automatically evaluate novel algorithms and applications. Exploring diverse scenarios of humans and robots interacting in simulation can improve understanding of the robotic system and avoid potentially costly failures in real-world settings. We formulate this problem as a quality diversity (QD) problem, where the goal is to discover diverse failure scenarios by simultaneously exploring both environments and human actions. We focus on the shared autonomy domain, where the robot attempts to infer the goal of a human operator, and adopt the QD algorithm MAP-Elites to generate scenarios for two published algorithms in this domain: shared autonomy via hindsight optimization and linear policy blending. Some of the generated scenarios confirm previous theoretical findings, while others are surprising and bring about a new understanding of state-of-the-art implementations. Our experiments show that MAP-Elites outperforms Monte-Carlo simulation and optimization based methods in effectively searching the scenario space, highlighting its promise for automatic evaluation of algorithms in human-robot interaction.
\end{abstract}

\IEEEpeerreviewmaketitle

\section{Introduction}


We present a method for automatically generating human-robot interaction (HRI) scenarios in shared autonomy. Consider as an example a manipulation task, where a user provides inputs to a robotic manipulator through a joystick, guiding the robot towards a desired goal, e.g., grasping a bottle on the table. The robot does not know the goal of the user in advance, but infers their desired goal in real-time by observing their inputs and assisting them by moving autonomously towards that goal. Performance of the algorithm is assessed by how fast the robot reaches the goal. However, different environments and human behaviors could cause the robot to fail, by picking the wrong object or colliding with obstacles.


\begin{figure}[!t]

\centering
\includegraphics[width=\linewidth]{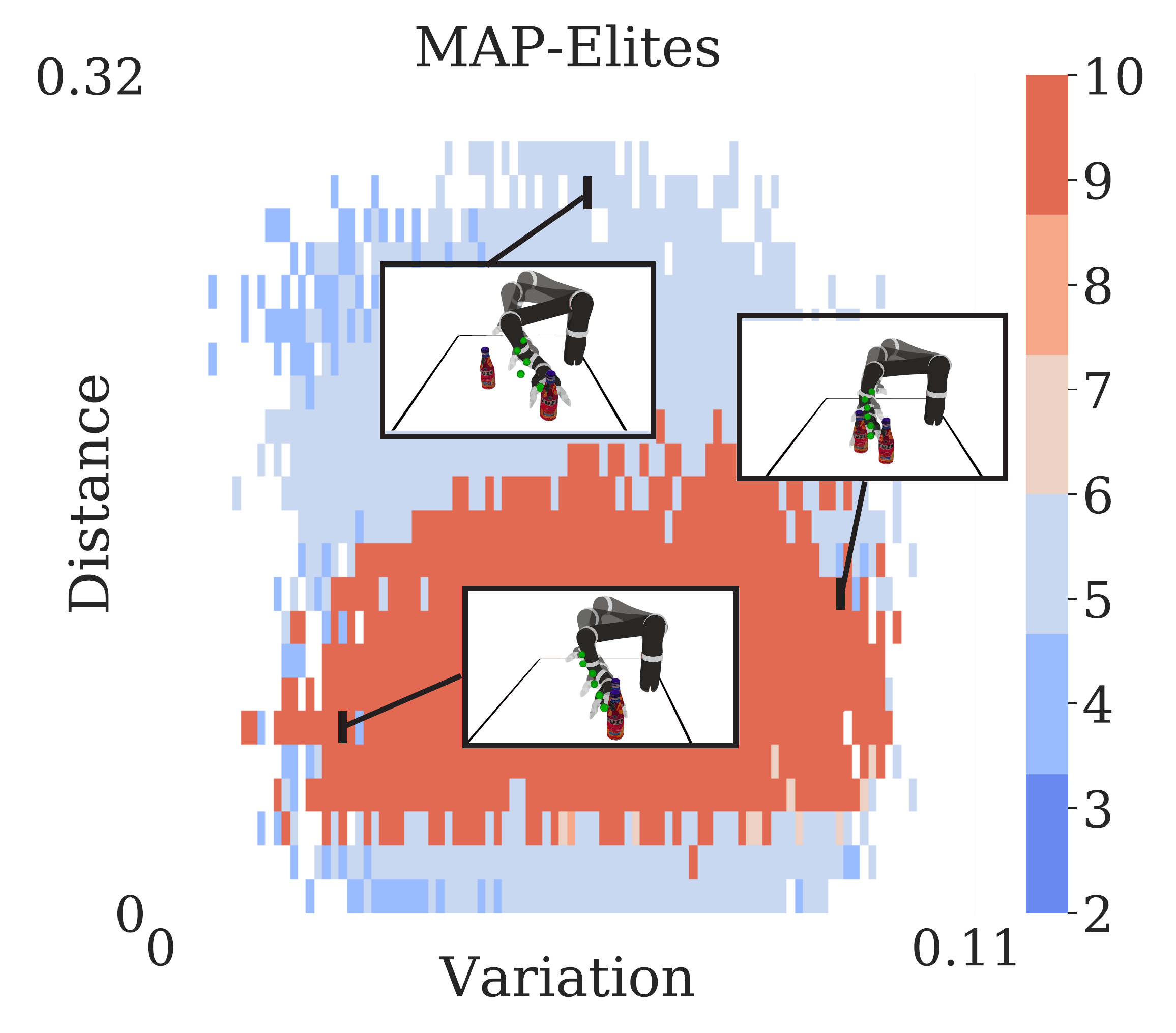}
\caption{An example archive of solutions returned by the quality diversity algorithm MAP-Elites. The solutions in red indicate scenarios where the robot fails to reach the desired user goal in a simulated shared autonomy manipulation task. The scenarios vary with the environment (y-axis: distance between the two candidate goals) and human inputs (x-axis: variation from optimal path).}

\label{fig:best}
\end{figure}

 Typically, such algorithms are evaluated with human subject experiments~\cite{thomaz2016computational}. While these experiments are fundamental in exploring and evaluating human-robot interactions and they can lead to exciting and unpredictable behaviors, they are often limited in the number of environments and human actions that they can cover. Testing an algorithm in simulation  with a \textit{diverse} range of scenarios can improve understanding of the system, inform the experimental setup of real-world studies, and help avoid potentially costly failures ``in the wild.'' 
 
 
 




One approach is to simulate agent behaviors by repeatedly sampling from models of human behavior and interaction protocols~\cite{steinfeld2009oz}. While this approach will show the \textit{expected} behavior of the system  given the pre-specified models, it is unlikely to reveal failure cases that are not captured by the models or are in the tails of the sampling distribution. Exhaustive search of human actions and environments is also computationally prohibitive given the continuous, high-dimensional space of all possible environments and human action sequences. 

Another approach is to formulate this as an optimization problem, where the goal is to find adversarial environments and human behaviors. But we are typically not interested in the maximally adversarial scenario, which is the single, global optimum of our optimization objective, since these scenarios are both easy to find and unlikely to occur in the real-world, e.g., the human moving the joystick of an assistive robotic arm consistently in the wrong direction. 

Instead, we are interested in answering questions of the form: how noisy can the human input be before the algorithm breaks? Or, in the aforementioned example task, how far apart do two candidate goals have to be for the robot to disambiguate the human intent?  





Our work makes the following contributions:\footnote{We include an overview video: \url{https://youtu.be/9P3qomydMWk}}

\textbf{1.}  We propose formulating the problem of generating human-robot interaction scenarios as a \textit{quality diversity} (QD) problem, where the goal is not to find a single, optimal solution, but a collection of high-quality solutions, in our case failure scenarios of the tested algorithm, across a range of measurable criteria, such as noise in human inputs and distance between objects. 

\textbf{2.}  We adopt the QD algorithm MAP-Elites, originally presented in~\cite{cully:nature15, mouret2015illuminating}, for the problem of scenario generation.  Focusing on the shared autonomy domain, where a robotic manipulator attempts to infer the user's goal based on their inputs, we show that MAP-Elites outperforms two baselines: standard Monte Carlo simulation (random search), where we uniformly sample the scenario parameters, and CMA-ES~\cite{hansen:cma16}, a state-of-the-art derivative-free optimization algorithm, in finding diverse scenarios that minimize the performance of the tested algorithm. We select to test the algorithm ``shared autonomy with hindsight optimization''~\cite{javdani2015hindsight}, since it has been widely used and we have found it to perform robustly in a range of different environments and tasks. Additionally, in hindsight optimization, inference and planning are tightly coupled, which makes testing particularly challenging; simply testing each individual component is not sufficient to reveal how the algorithm will perform.

 

\textbf{3.}  We show that Monte Carlo simulation does not perform well because of \textit{behavior space distortion}: sampling directly from the space of environments and human actions covers only a small region in the space of measurable aspects (behavioral characteristics). For example, uniformly sampling object locations (scenario parameters) results in a non-uniform distribution of their distances (behavioral characteristic) with a very small variance near the mean. On the other hand, MAP-Elites focuses on exploring the space of the behavioral characteristics by retaining an archive of high-performing solutions in that space and perturbing existing solutions with small variations. Therefore, MAP-Elites performs a type of simultaneous search guided by the behavioral characteristics, where solutions in the archive are used to generate future candidate solutions~\cite{mouret2015illuminating}.

\textbf{4.} We analyze the failure cases and we show that they result from specific aspects of the implementation of the tested algorithm, rather than being artifacts of the simulation environment. We use the same approach to contrast the performance of hindsight optimization with that of linear policy blending~\cite{dragan2012formalizing} and generate a diverse range of scenarios that 
confirm previous theoretical findings~\cite{trautman2015assistive}. The generated scenarios transfer to the real-world; we reproduce some of the automatically discovered scenarios on a real robot with human inputs. While some of the scenarios are expected, e.g., the robot approaches the wrong goal if the human provides very noisy inputs, others are surprising, e.g, the robot never reaches the desired goal even for a nearly optimal user if the two objects are aligned in column formation in front of the robot (Fig.~\ref{fig:best})!


QD algorithms treat the algorithm being tested as a ``black box'', without any knowledge of its implementation, which makes them applicable to multiple domains. 
Overall, we are excited about the potential of QD to facilitate understanding of complex HRI systems, opening up a number of scientific challenges and opportunities to be explored in the future. 








\section{Problem Statement} \label{sec:problem}



Given a shared autonomy system where a robot interacts with a human, our goal is to generate scenarios that minimize performance of the system, while ensuring that the generated scenarios cover a range of prespecified measures.

We let $R$ be a single robot interacting with a single human $H$. We assume a function $G_H$ that generates human inputs, a function $G_E$ that generates an environment, and an HRI algorithm $G_R$ that generates actions for the robot. The human input generator is parameterized by $\theta \in \mathbb{R}^{n_{\theta}}$, where $n_{\theta}$ is the dimensionality of the parameter space, while the environment generator is parameterized by $\phi \in \mathbb{R}^{n_{\phi}}$. We define a \textit{scenario} as the tuple $(\theta, \phi)$. 

In shared autonomy, $G_E(\phi)$ generates an initial environment (and robot) state $x_E$. The human observes $x_E$ and provides inputs to the system $u_H = G_H(x_E, \theta)$ through some type of interface. The robot observes $x_E$ and the human input $u_H$ and takes an action 
 $u_R = G_R(x_E, u_H)$. The state  changes with dynamics: $\dot{x}_E = h(x_E, u_R)$. H and R interact for a time horizon $T$, or until they reach a final state $x_f \in X_E$. 
 
 To evaluate a scenario, we assume a function \\ $f(x_E^{0..T}, u_R^{0..T},u_H^{0..T}) \rightarrow \mathbb{R}$ that maps the state and action history to a real number.  We call this an \textit{assessment} function, which measures the performance of the robotic system. We also assume $M$ user-defined functions, $b_i(x_E^{0..T}, u_R^{0..T},u_H^{0..T})\rightarrow \mathbb{R},~i\in[M]$. These functions measure aspects of generated scenarios that should vary, e.g., noise in human inputs or distance between obstacles. We call these functions \textit{behavior characteristics} (BCs), which induce a Cartesian space called a \textit{behavior space}.

Given the parameterization of the environment and human input generators, we can map a value assignment of the parameters $(\theta, \phi)$ to a state and action  history $(x_E^{0..T},u_R^{0..T},u_H^{0..T})$ and therefore to an assessment $f(\theta, \phi)$ and a set of  BCs $b(\theta,\phi)$. We assume that the behavior space is partitioned into N cells, which form an \textit{archive} of scenarios, and we let $(\theta_i, \phi_i)$ be the parameters of the scenario occupying cell $i \in [N]$.

The objective of our scenario generator is to fill in as many cells of the archive as possible with scenarios of high assessment $f$: \footnote{We note that the assessment function could be any performance metric of interest, such as time to completion or minimum robot's distance to obstacles. Additionally, while in this work we focus on minimizing performance, we could instead search for scenarios that maximize performance, or that achieve performance that matches a desired value. We leave this for future work.
}
\begin{equation}
  \mathcal{M}(\theta_1, \phi_1, ... , \theta_N, \phi_N) =  \max \sum_{i=1}^N f(\theta_i, \phi_i)
\label{eq:objective}
\end{equation}

\section{Background} \label{sec:background}
\noindent\textbf{Automatic Scenario Generation }
Automatically generating scenarios is a long standing problem in human training~\citep{hofer1998automated}, with the core challenge being the generation of \textit{realistic} scenarios~\citep{martin:pcg10}. Previous work~\citep{zook2012automated} has shown that optimization methods can be applied to generate scenarios by maximizing a scenario quality metric. 

Scenario generation has been applied extensively to evaluating autonomous vehicles~\citep{arnold:safecomp13,mullins:av18, abey:av19, rocklage:av17, gambi:av19,sadigh2019verifying}. Contrary to model-checking and formal methods~\cite{choi2013model,o2014automatic}, which require a model describing the system's performance such as a finite-state machine~\cite{meinke2015learning} or process algebras~\cite{o2014automatic},
 black-box approaches do not require access to a model. Most relevant are black-box falsification methods~\cite{deshmukh2017testing,zhao2003generating,kapinski2016simulation,dreossi2019verifai} that attempt to find an input trace that minimizes the performance of the tested system. Rather than searching for a single global optimum~\cite{deshmukh2017testing, deshmukh2015stochastic,sadigh2019verifying}, or attempting to maximize coverage of the space of scenario parameters~\cite{zhao2003generating} or of performance boundary regions~\cite{mullins:av18}, we propose a quality diversity approach where we optimize an archive formed by a set of  behavioral characteristics, with a focus on the shared autonomy domain. This allows us to \textit{simultaneously} search for human-robot interaction scenarios that minimize the performance of the system over a range of measurable criteria, e.g., over a range of variation in human inputs and distance between goal objects.

Finally, scenario generation is closely related to the problem of generating video game levels in procedural content generation (PCG)~\citep{hendrikx2013procedural,shaker:book16}. An approach gaining popularity is procedural content generation through quality diversity (PCG-QD)~\citep{gravina2019procedural}, which leverages QD algorithms to drive the search for interesting and diverse content.

\noindent\textbf{Quality Diversity and MAP-Elites. }
QD algorithms differ from pure optimization methods, in that they do not attempt to find a single optimal solution, but a collection of good solutions that differ across specified dimensions of interest. For example, QD algorithms have generated video game levels of varying number of enemies or tile distributions~\cite{khalifa2019intentional, fontaine2020illuminating}, and objects of varying shape complexity and grasp difficulty~\cite{morrison2020egad}. 

\mbox{MAP-Elites} \citep{mouret2015illuminating,cully:nature15} is a popular QD algorithm that searches along a set of explicitly defined attributes called \textit{behavior characteristics} (BCs), which induce a Cartesian space called a \textit{behavior space}. The behavior space is tessellated into uniformly spaced grid cells. In each cell, the algorithm maintains the highest performing solution, which is called as \textit{elite}. The collection of elites returned by the algorithm forms an \textit{archive} of solutions. 


 MAP-Elites  populates the archive by first randomly sampling a population of solutions, and then selecting the elites -- which are the top performing solutions in each cell of the behavior space -- at random and perturbing them with small variations. The objective of the algorithm is two-fold: maximize the number of filled cells (coverage) and maximize the quality of the elite in each cell. Recent algorithms have focused on how the behavior space is tessellated~\citep{smith:ppsn16,fontaine:gecco19}, as well as how each elite is perturbed~\cite{vassiliades:gecco18}. Recent work~\cite{fontaine2021differentiable} has also shown that, when the objective function and behavior characteristics are first-order differentiable, MAP-Elites via a Gradient Arborescence (MEGA) can result in significant improvements in search efficiency. 

By retaining an archive of high-performing solutions and perturbing existing solutions with small variations, \mbox{MAP-Elites} simultaneously optimizes every region of the archive, using existing solutions as ``stepping stones'' to find new solutions. Previous work has shown that \mbox{MAP-Elites} variants~\cite{mouret2020quality} and surrogate models~\cite{gaier2017data} outperform independent single-objective constrained optimizations for each cell with \mbox{CMA-ES}, with the same total budget of evaluations.

 \noindent\textbf{Coverage-Driven Testing in HRI.} Previous work~\cite{araiza2015coverage,araiza2016systematic} explored test generation in human-robot interaction using Coverage-Driven Verification (CDV), emulating techniques used in functional verification of hardware designs. Human action sequences were randomly generated in advance and with a model-based generator which modeled the interaction with Probabilistic-Timed Automata. Instead, we focus on online scenario generation by searching over a set of scenario parameters; the generator itself is agnostic to the underlying HRI algorithm and human model. Previous work~\cite{araiza2016intelligent} has also used Q-learning to generate plans for an agent in order to maximize coverage. Our focus is both on coverage and quality of generating scenarios, with respect to a prespecified set of behavioral characteristics that we want to cover. In contrast to previous studies that simulate human actions, \textit{we jointly search for environments and human/agent behaviors.}

\noindent\textbf{Shared Autonomy.}
Shared autonomy (also: shared control, assistive teleoperation) combines human teleoperation of a robot with intelligent robotic assistance. The method has been applied in the control of robotic arms~\cite{javdani2018shared,dragan2012formalizing,nikolaidis2017mutualadaptation,Muelling2017,herlant2016assistive,gopinath2016human, jain2019probabilistic,jeon2020shared, losey2019controlling,rakita2019shared,rakita2018shared}, the flight of UAVs~\cite{reddy2018shared,gillula2011applications,lam2009artificial}, and robot-assisted surgery~\cite{li2003recognition,ren2008dynamic}. Shared autonomy has been implemented through a variety of interfaces, such as whole body motions~\cite{dragan2012formalizing}, natural language \cite{doshi2007efficient}, laser pointers \cite{veras2009scaled}, brain-computer interfaces \cite{Muelling2017}, body-machine interfaces~\cite{jain2015assistive}, and eye gaze~\cite{bien2004,javdani2018shared}. A shared autonomy system first predicts the human's goal, often through machine learning methods trained from human demonstrations \cite{hauser13,koppula16,wang13}, and then provides assistance, which often involves blending the user's input with the robot assistance to achieve the predicted goal \cite{dragan2012formalizing,fagg04,kofman05}. Assistance can provide task-dependent guidance \cite{aarno2005adaptive},  manipulation of objects~\cite{jeon2020shared}, or mode switches \cite{herlant2016assistive}.

\noindent\textbf{Shared Autonomy via Hindsight Optimization.}
In shared autonomy via hindsight optimization~\cite{javdani2015hindsight} assistance blends user input and robot control based on the confidence of the robot's goal prediction. The problem is formulated as a Partially Observable Markov Decision Process (POMDP), wherein the user's goal is a latent variable. The system models the user as an approximately optimal stochastic controller, which provides inputs so that the robot reaches the goal as fast as possible. The system treats the user's inputs as observations to update a distribution over the user's goal, and assists the user by minimizing the expected cost to go -- estimated using the distance to goal -- for that distribution. Since solving a POMDP exactly is intractable, the system uses the hindsight optimization (QMDP) approximation~\cite{littman1995learning}. The system was shown to achieve significant improvements in efficiency of manipulation tasks in an object-grasping task~\cite{javdani2015hindsight} and more recently in a feeding task~\cite{javdani2018shared}. We empirically found this algorithm to perform robustly in a range of different environments, which motivates a systematic approach for testing. We refer to this algorithm simply as \textit{hindsight optimization}.

\section{Scenario Generation with MAP-Elites}

Algorithm~\ref{alg:map-elites} shows the MAP-Elites algorithm from~\cite{mouret2015illuminating,cully:nature15}, adapted for scenario generation. The algorithm takes as input a function $G_H$ parameterized by $\theta$ that generates human inputs, a function $G_E$ parameterized by $\phi$ that generates environments, and an HRI algorithm $G_R$ that generates actions for the robot. The algorithm searches for scenarios $\theta, \phi$ of high assessment values  $f$ that fill in the archive $\mathcal{P}$. 

For each scenario $(\theta, \phi)$, MAP-Elites instantiates the generator functions $G_H$ and $G_E$. For instance, $\phi$ could be a vector of objects positions, and $\theta$ could be a vector of waypoints representing a trajectory of human inputs, or parameters of a human policy. 

When a scenario is generated, MAP-Elites executes the scenario in a simulated environment and estimates the assessment function $f$ and the BCs $b$. MAP-Elites then updates the archive  if (1) the cell corresponding to the BCs $\mathcal{X}[b]$ is empty, or (2) the existing scenario (elite) in $\mathcal{X}[b]$ has a smaller assessment function (lower quality) than the new scenario.  This allows populating the archive to maximize coverage as well as improving the quality of existing scenarios.

For the first $N_{init}$ iterations, the algorithm generates scenarios $\theta, \phi$ by randomly sampling from the parameter space. These sampled parameters seed the archive with an initial set of scenarios. After the first $N_{init}$ iterations, MAP-Elites selects a scenario uniformly at random from the archive and perturbs it with a small variation. This allows for better exploration of the archive, compared to random search, as we show in section~\ref{subsec:Analysis}.

We note that, while our experiments focus on the shared autonomy domain, the proposed scenario generation method is general and can be applied to multiple HRI domains.

\begin{algorithm}[t!]
 \caption{Scenario Generation with MAP-Elites}
\label{algorithm: known context}
\begin{algorithmic}
\STATE\textbf{Input:} Human input generator $G_H$, environment generator $G_E$, HRI algorithm $G_R$, variations
$\sigma_{\theta},\sigma_{\phi}$
\STATE\textbf{Initialize:} Scenarios in archive $\mathcal{X}\leftarrow \emptyset$, assessments $\mathcal{F}\leftarrow \emptyset$

\FOR{$t=1,\ldots,N $}
\IF{$t < N_{init}$}
\STATE Generate scenario $\theta, \phi = \mathrm{random\_generation()}$ 
\ELSE
\STATE Select elite $\theta', \phi' = \mathrm{random\_selection}(\mathcal{X})$
\STATE Sample $\theta \sim N(\theta', \sigma_\theta)$ 
\STATE Sample $\phi \sim N(\phi', \sigma_\phi)$ 

\ENDIF

\STATE Instantiate $G_H^{\theta} = G_H(\theta)$
\STATE Instantiate $G_E^{\phi} = G_E(\phi)$
\STATE Compute $f = \textrm{assessment}(G_H^{\theta},G_E^{\phi}, G_R)$
\STATE Compute $b = \textrm{behaviors}(G_H^{\theta},G_E^{\phi}, G_R)$
 \IF{$\mathcal{F}[b]= \emptyset$ or $\mathcal{F}[b] < f$ } 
 \STATE Update archive $\mathcal{X}[b] \leftarrow (\theta, \phi),\mathcal{F}[b] \leftarrow f$   
 \ENDIF
\ENDFOR
\end{algorithmic}
\label{alg:map-elites}
\end{algorithm}

\section{Generating Scenarios in Shared Autonomy} \label{sec:limiting}

We focus on a shared autonomy manipulation task, where a human user teleoperates a robotic manipulator through a joystick interface. The robot runs a hindsight optimization shared autonomy algorithm~\cite{javdani2015hindsight}, which uses the user's input to infer the object the user wants the robot to grasp, and assists the user by moving autonomously towards that goal.  

\subsection{Scenario Parameters} \label{subsec:parameters}
Following the specification of section~\ref{sec:problem}, we define a human input generator $G_H$ parameterized by $\theta$ and an environment generator $G_E$ parameterized by $\phi$.

\noindent\textbf{Environment Generator:} The environment generator $G_E$ takes as input the 2D positions $g_i$ of $n$ goal objects (bottles), so that $\phi = (g^1, ..., g^n)$, and places them on top of a table. We specify the range of the coordinates $g_x \in [0, 0.25]$ (in meters), $g_y \in [0, 0.2]$ so that the goals are always reachable by the robot's end-effector. We position the robotic manipulator to face the objects (Fig.~\ref{fig:elites}).

\noindent\textbf{Human Input Generator:} Our pilot studies with the shared autonomy system have shown that user inputs are typically not of constant magnitude. Instead, inputs spike when users wish to ``correct'' the robot's path, and decrease in magnitude afterwards when the robot takes over. 

Therefore, we specify the human input generator $G_H$, so that it generates a set of equidistant waypoints in Cartesian space forming a straight line that starts from the initial position of the robot's end-effector and ends at the desired goal of the user. At each timestep, the generator takes as input the current state of the robot (and the environnment) $x_E$, and provides a translational velocity command $u_H$ for the robot's end-effector towards the next waypoint, proportional to the distance to that waypoint. 

We allow for noise in the waypoints by adding a disturbance $d \in [-0.05, 0.05]$ for each of the intermediate waypoints in the horizontal direction (x-axis in Fig.~\ref{fig:elites}). We selected $m=5$ intermediate waypoints, and specified the human input parameter $\theta$ as a vector of disturbances, so that $\theta = (d_1, ... , d_5)$. We note that this is only one way, out of many, of simulating the human inputs.

\noindent\textbf{HRI Algorithm and Simulation Environment:} We use the publicly available implementation of the hindsight optimization algorithm~\cite{ada_code}, which runs on the OpenRAVE~\cite{diankov2008openrave} simulation environment. Experiments were conducted with a Gen2 Lightweight manipulator. For each goal object we assume one target grasp location, on the side of the object that is facing the robot.


\subsection{Assessment Function.}  The assessment function $f$ represents the quality of a scenario. We evaluate a scenario by simulating it until the robot's end-effector reaches the user's goal, or when the maximum time (10 $s$) has elapsed. We use as an assessment function time to completion, where longer times represent higher scenario quality, since we wish to discover scenarios that \textit{minimize} performance. 

\subsection{Behavioral Characteristics.} 
We wish to generate scenarios that show the limits of the shared autonomy system: how noisy can the human be without the system failing to reach the desired goal? How does distance between candidate goals affect the system's performance? Intuitively, noisier human inputs and smaller distances between goals would make the inference of the user's goal harder and thus make the system more likely to fail. 

These dimensions of interest are the behavioral characteristics (BC) $b$: attributes that we wish to obtain coverage for. We explore the following BCs:


\noindent\textbf{Distance Between Goals:} How far apart the human goal is from other candidate goals in a scenario plays an important role in disambiguating the human intent when the robot runs the hindsight optimization algorithm. The reason is that the implementation of the algorithm models the human user as minimizing a cost function proportional to the distance between the robot and the desired goal. The framework then infers the user's goal by using the user inputs as observations; the more unambiguous the user input, the more accurate the inference of the system.  Therefore, we expect that the further away the human goal object $g_H$ is from the nearest goal $g_N$, the better the system will perform. We define this BC as: 
\begin{equation}
BC1 = ||g_H - g_N||_2
\end{equation}

Given the range of the goal coordinates, the range of this BC is $[0, 0.32]$. In practice, there will be always a minimum distance between two goal objects because of collisions, but this does not affect our search, since we can ignore cases where the objects collide. We partitioned this behavior space to 25 uniformly spaced intervals.

\noindent\textbf{Human Variation:} 
We expect noise in the human inputs to affect the robot's inference of the user's goal and thus the system's performance. 
We capture variation from the optimal path using the root sum of the squares of the disturbances $d_i$ applied to the $m$ intermediate waypoints.
\begin{equation}
BC2 = \sqrt{\sum_{i=1}^m d_i^2}
\end{equation}
 A value of 0 indicates a straight line to the goal. Since we have $d_i \in [-0.05,0.05]$ (section~\ref{subsec:parameters}), the range of this BC is $[0,0.11]$. We partitioned this behavior space to 100 uniformly spaced intervals.


\noindent\textbf{Human Rationality:} If we interpret the user's actions using a bounded rationality model~\cite{baker2007goal,fisac2018probabilistically}, we can explain deviations from the optimal trajectory of human inputs as a result of how ``rational'' or ``irratonal'' the user is.\footnote{We note that we use the human rationality model as one way, out of many, to \textit{interpret} human inputs and not to as a way to  \textit{generate} inputs. Human inputs can be generated with any generator model. In this paper, we generate human inputs with the deterministic model described in section~\ref{subsec:parameters}. We discuss extensions to stochastic human models in section~\ref{sec:discussion}.}

 Formally, we let $x_R$ be the 3D position of the robot's end-effector and $u_H$ be the velocity controlled by the user in Cartesian space. We model the user as following  Boltzmann policy $\pi_H \mapsto P(u_H|x_R,g_H, \beta)$, where $\beta$ is the rationality coefficient -- also interpreted as the expertise~\cite{jeon2020shared}-- of the user and $Q_{g_H}$ is the value function from $x_R$ to the goal $g_H$.
\begin{equation}
P(u_H|x_R,g_H, \beta)  \propto 
e^{- \beta Q_{g_H}(x_R, u_H)}
\label{eqn:human}
\end{equation}

Let $Q_{g_H} = -||u_H||_2 - ||x_R + u_H - g_H||_2$~\cite{fisac2018probabilistically}, so that the user minimizes the distance to the goal. Observe that if $\beta \rightarrow \infty$, the human is rational, providing velocities exactly in the direction to the goal. If $\beta \rightarrow 0$, the human is random, choosing actions uniformly. 

We can estimate the user's rationality, given their inputs, with Bayesian inference~\cite{fisac2018probabilistically}: 

\begin{equation}
P(\beta|x_R,g_H, u_H)  \propto 
P(u_H|x_R,g_H, \beta) P(\beta)
\label{eqn:human}
\end{equation}

Since the human inputs change at each waypoint (section~\ref{subsec:parameters}), we perform $m+1$ updates, at the starting position and at each intermediate waypoint, on a finite set of discrete values $\beta$. Following previous work~\cite{jeon2020shared}, we set the rationality range  $\beta \in [0,1000]$. We then choose as behavioral characteristic the value with the maximum a posteriori probability at the end of the task:

\begin{equation}
BC3 = \argmax P(\beta|x^{0..T}_R, g_H, u^{0..T}_H)
\end{equation}

We partitioned the space to 101 uniformly spaced intervals.

\begin{table*}[t]
\centering
\resizebox{.8\linewidth}{!}{
\begin{tabular}{l|cc|cc|cc}
\hline
             & \multicolumn{2}{l|}{BC1 \& BC3, 2 goals}          & \multicolumn{2}{l|}{BC1 \& BC2, 2 goals} & \multicolumn{2}{l}{BC1 \& BC2, 3 goals}  \ \\ 
    \toprule
Algorithm     & Coverage & QD-Score & Coverage & QD-Score & Coverage & QD-score\\
    \midrule
Random  &  22.3\% & 3464 & 48.4\% &7782     & 41.9\% &  7586\\
CMA-ES   & 24.8\% & 4540& 38.9\% &  7422 & 34.5\% & 7265\\
MAP-Elites    & \textbf{62.8\%} & \textbf{10128} &  \textbf{63.0\%} &  \textbf{11216} &  \textbf{57.4\%} &  \textbf{11204}\\
  \bottomrule
\end{tabular}
}
\caption{Results: Percentage of cells covered (coverage) and QD-Score after 10,000 evaluations,  averaged over 5 trials.}
\label{tab:results}
\end{table*}


\begin{figure*}[t!]
    \centering
        \includegraphics[width=1.0\textwidth]{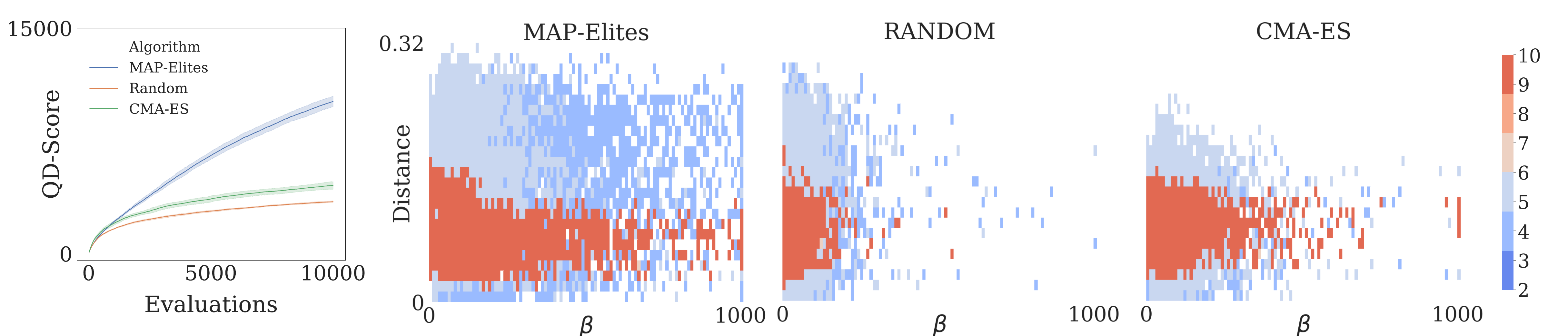}
      \includegraphics[width=1.0\textwidth]{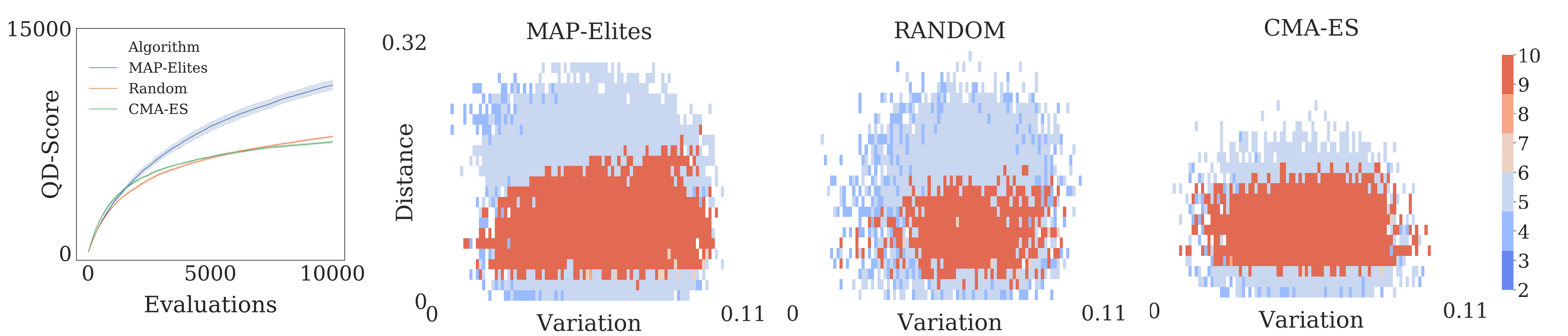}
\caption{QD-Scores over evaluations (generated scenarios) and example archives returned by the three algorithms for the first two behavior spaces of Table~\ref{tab:results}. The colors of the cells in the archives represent time to task completion in seconds.}
\label{fig:maps}
\end{figure*}



\section{Experiments}

We compare different search algorithms in their ability to find diverse and high-quality scenarios in different behavior spaces.

\subsection{Independent Variables}
The experiment has two independent variables, the \textit{behavior space} and the \textit{search algorithm}.

\textit{Behavior Space}: (1) Distance between $n=2$ goal objects (BC1) and human rationality (BC3), (2) Distance between $n=2$ goal objects (BC1) and human variation (BC2), and (3) Distance between human goal and nearest goal for $n=3$ goals (BC1) and human variation (BC2).\footnote{We note that the behavior spaces can be more than two-dimensional, e.g, we could specify a space with all three BCs. We include only 2D spaces since they are easier to visualize and inspect.}


\textit{Search Algorithm}: We evaluate three different search methods: MAP-Elites, CMA-ES and random search. The Covariance Matrix Adaptation Evolution Strategy (\mbox{CMA-ES}) is one of the most competitive derivative-free optimizers for single-objective optimization of continuous spaces (see~\citep{hansen:cma16,hansen2009benchmarking}) and it is commonly used for falsification of cyber-physical systems~\cite{deshmukh2015stochastic,zhang2018time}. In random search we use Monte Carlo simulation where scenario parameters are sampled uniformly within their prespecified ranges.

We implemented a multi-processing system on an AMD  Ryzen Threadripper 64-core (128 threads) processor, as a master search node and multiple worker nodes running separate OpenRAVE processes in parallel, which enables simultaneous evaluation of many scenarios. Random search and MAP-Elites run asynchronously on the master search node, while \mbox{CMA-ES} synchronizes before each covariance matrix update. We generated 10,000 scenarios per trial, and ran 45 trials, 5 for each algorithm and behavior space. One trial parallelized into 100 threads lasted approximately 20 minutes. 

\subsection{Algorithm Tuning}
MAP-Elites first samples uniformly the space of scenario parameters $\theta, \phi$ within their prespecified ranges for an initial population of $100$ scenarios (Algorithm~\ref{alg:map-elites}). The algorithm then 
randomly perturbs the elites (scenarios from the archive) with Gaussian noise scaled by a $\sigma$ parameter. The two scenario parameters, position of goal objects $\phi$ and human waypoints $\theta$, are on different scales, thus we specified a different $\sigma$  for each: $\sigma_\phi = 0.01, \sigma_\theta = 0.005$. 

To generate the scenarios for random search, we 
uniformly sample scenario parameters within their prespecified ranges, a method identical to generating the initial population of MAP-Elites. 

For CMA-ES, we selected a population of $\lambda = 12$ following the recommended setting from~\cite{hansen:cma16}. To encourage exploration, we used the bi-population variant of CMA-ES with restart rules~\cite{auger2005restart,hansen2009benchmarking}, where the population doubles after each restart, and we selected a large step size, $\sigma = 0.05$. Since the two search parameters are in different scales, we initialized the diagonal elements of the covariance matrix $C$, so that $c_{ii} = 1.0, i \in [2n]$ and $c_{ii} = 0.5, i \in \{2n+1, ..., 2n+m\}$, with $2n$ and $m$ the dimensionality of the goal object and human input parameter spaces respectively. 

Both CMA-ES and MAP-Elites may sample scenario parameters that do not fall inside their prespecified ranges. Following recent empirical results on bound constraint handling~\cite{biedrzycki2020handling}, we adopted a resampling strategy, where new scenarios are resampled until they fall within the prespecified range.   
\subsection{Measures} 
We wish to measure both the diversity and the quality of scenarios returned by each algorithm. These are combined by the QD-Score metric~\cite{pugh2015confronting}, which is defined as the sum of $f$ values of all elites in the archive (Eq.~\ref{eq:objective} in section~\ref{sec:problem}). Empty cells have 0 $f$ value. Therefore, QD-score is positively affected by both the coverage of the archive (the number of occupied cells in the archive divided by the total number of cells) and the assessment of the occupied cells. Similarly to previous work~\cite{fontaine:gecco20}, we compute the QD-Score of CMA-ES and random search for comparison purposes by calculating the behavioral characteristics for each scenario and populating a pseudo-archive. We also include the coverage score as an additional metric of diversity.


\subsection{Hypothesis}
\textit{We hypothesize that MAP-Elites will result in larger QD-Score  and coverage than both CMA-ES and random search.}

Previous work~\cite{fontaine:gecco19,fontaine:gecco20} has shown that behavior spaces are typically distorted: uniformly sampling the search parameter space results in samples concentrated in small areas of the behavior space. Therefore, we expect random search to have small coverage of the behavior space. Additionally, since random search ignores the assessment function $f$, we expect the quality of the found scenarios in the archive to be low.

CMA-ES moves with a single large population that has global optimization properties. Therefore, we expect it to concentrate in regions of high-quality scenarios, rather than explore the archive. On the other hand, MAP-Elites both expands the archive and maximizes the quality of the scenarios within each cell.

\begin{figure*}[t!]
    \centering
    \begin{tabular}{ccc}
    \begin{subfigure}[t]{.25\textwidth}
        \centering
        \includegraphics[width=\textwidth]{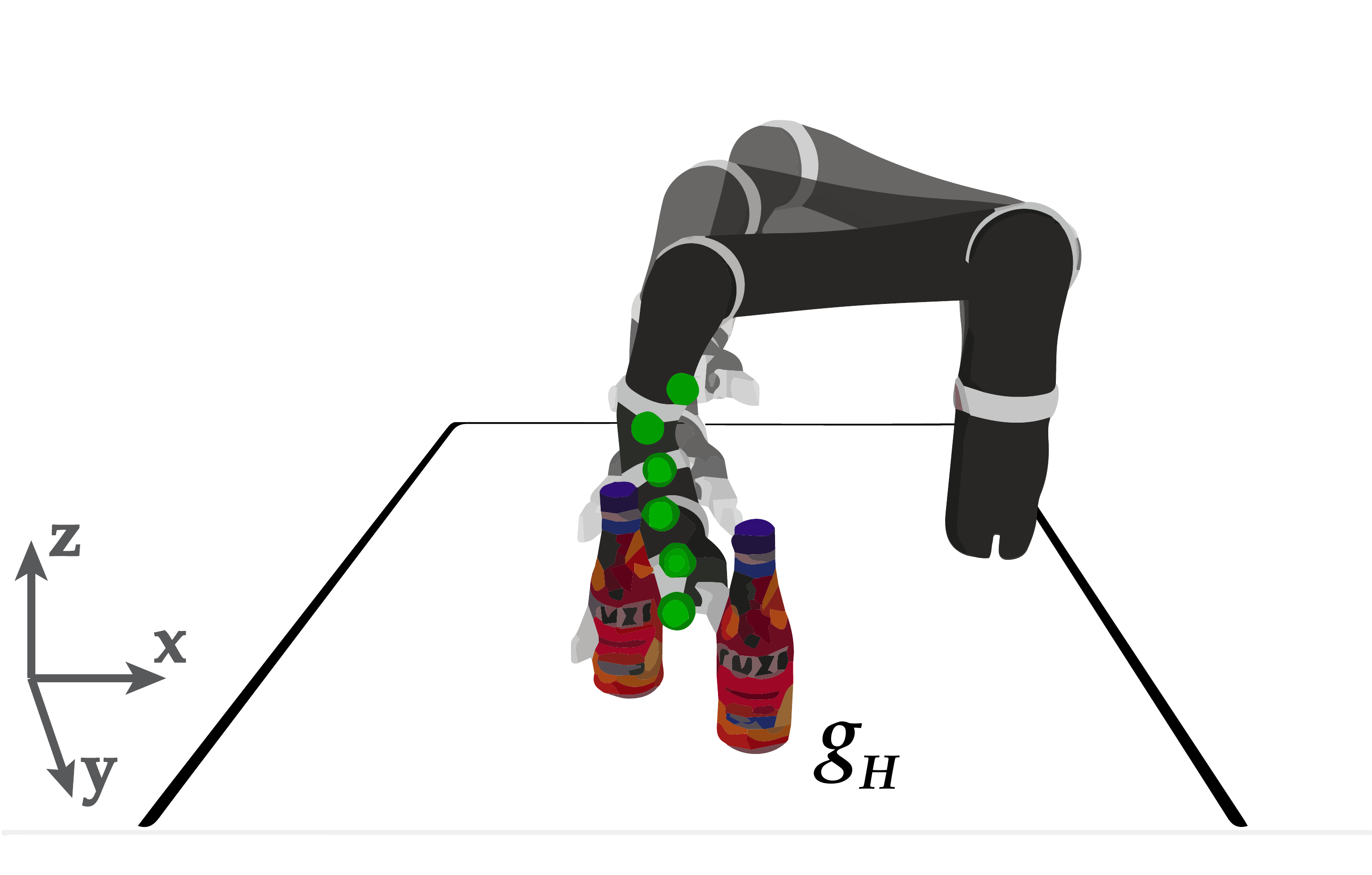}
    \end{subfigure} & 
    \begin{subfigure}[t]{.25\textwidth}
        \centering
        \includegraphics[width=\textwidth]{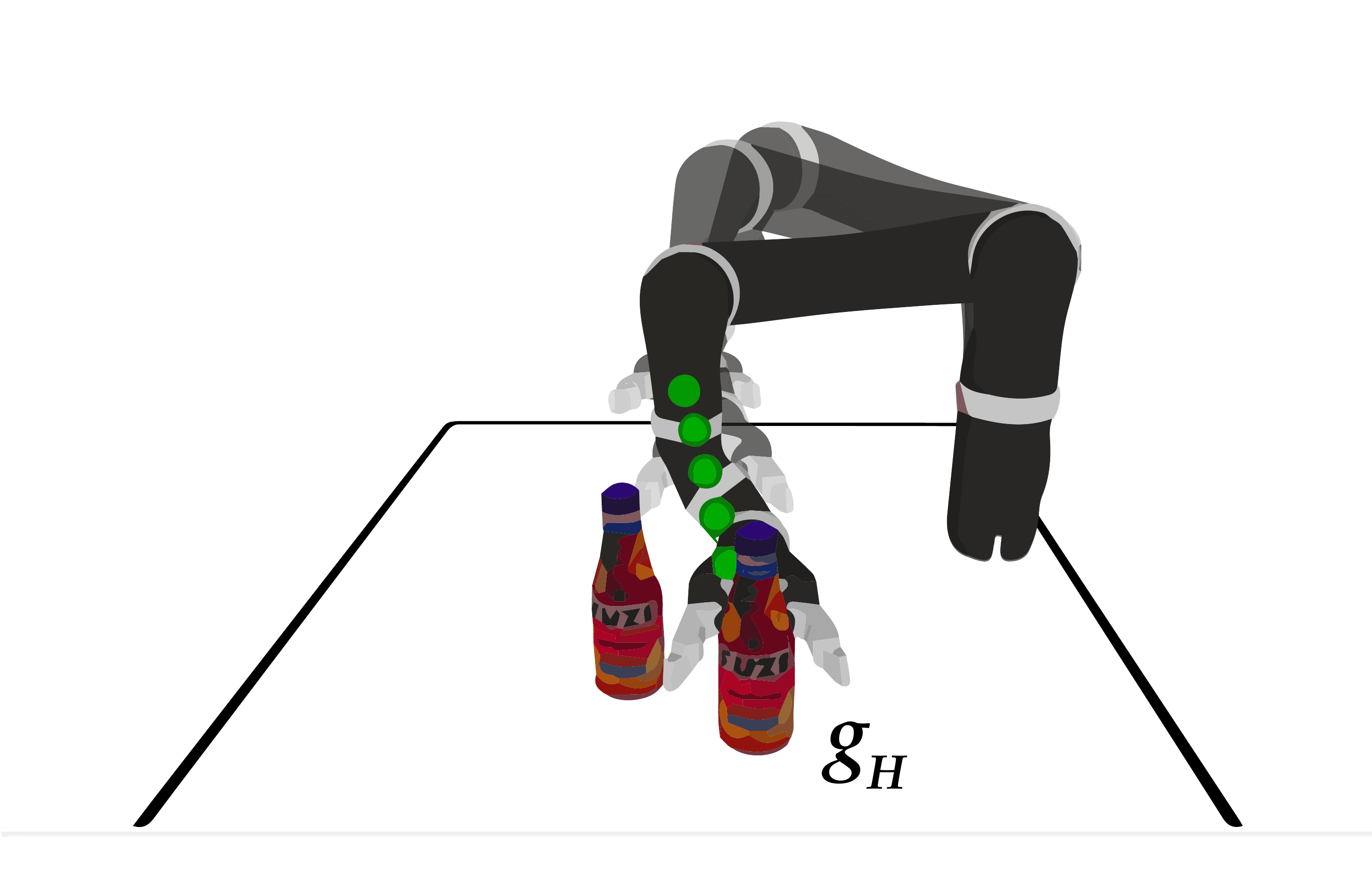}
        \end{subfigure}&
      \begin{subfigure}[t]{.25\textwidth}
        \centering
        \includegraphics[width=\textwidth]{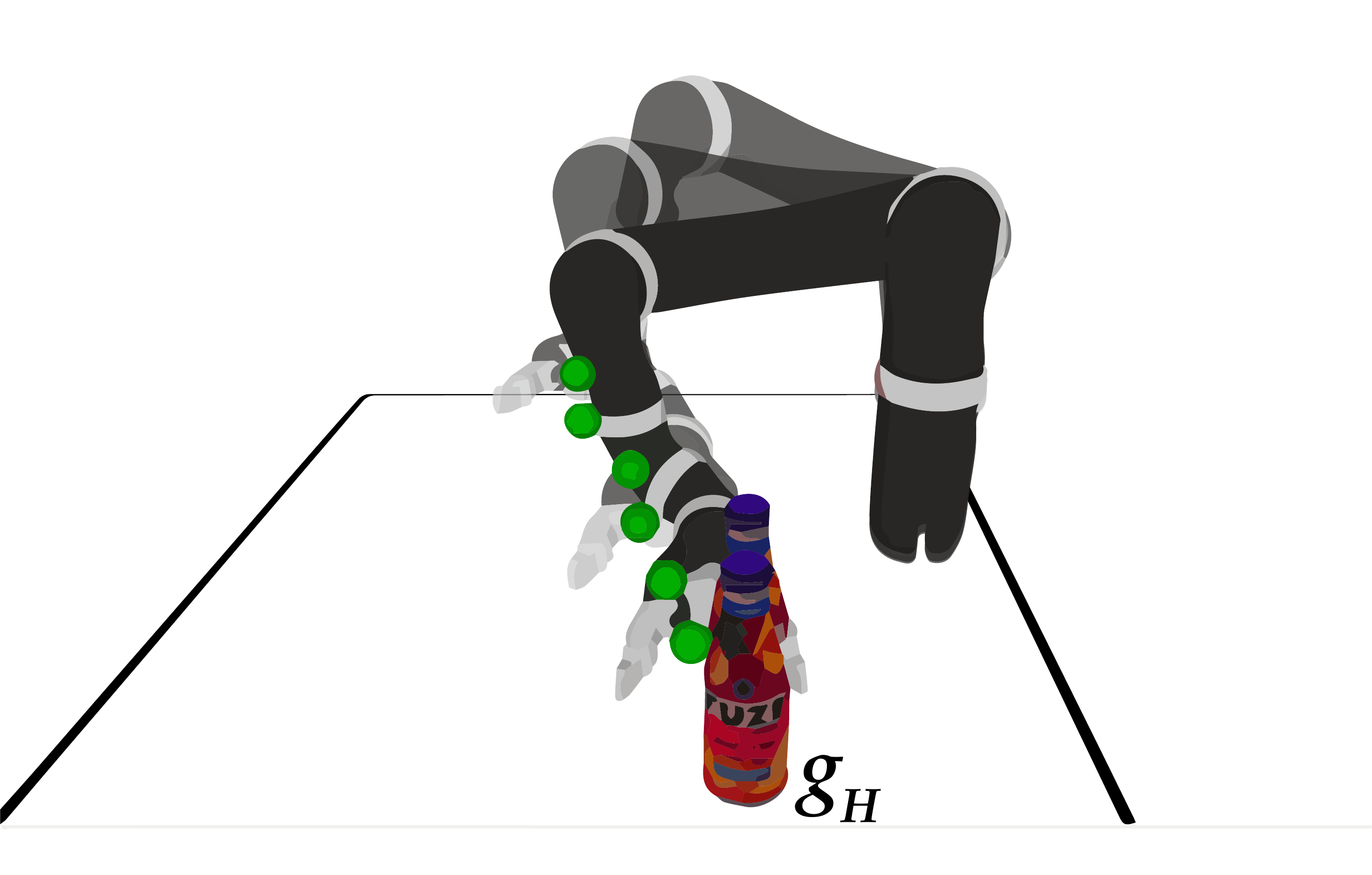}
    \end{subfigure}  \\ 
   
    \end{tabular}
\caption{(Left) The robot fails to reach the user's goal $g_H$ because of the large deviation in human inputs from the optimal path. The waypoints of the human inputs are indicated with green color. (Center) We show for comparison how the robot would act if human deviation was 0 (optimal human). (Right) The robot fails to reach the user's goal $g_H$ (bottle furthest away from the robot), even though the human provides a near optimal input trajectory.}
\label{fig:elites}
\end{figure*}


   \begin{figure}[t!]
    \centering
    \includegraphics[width=0.8\columnwidth]{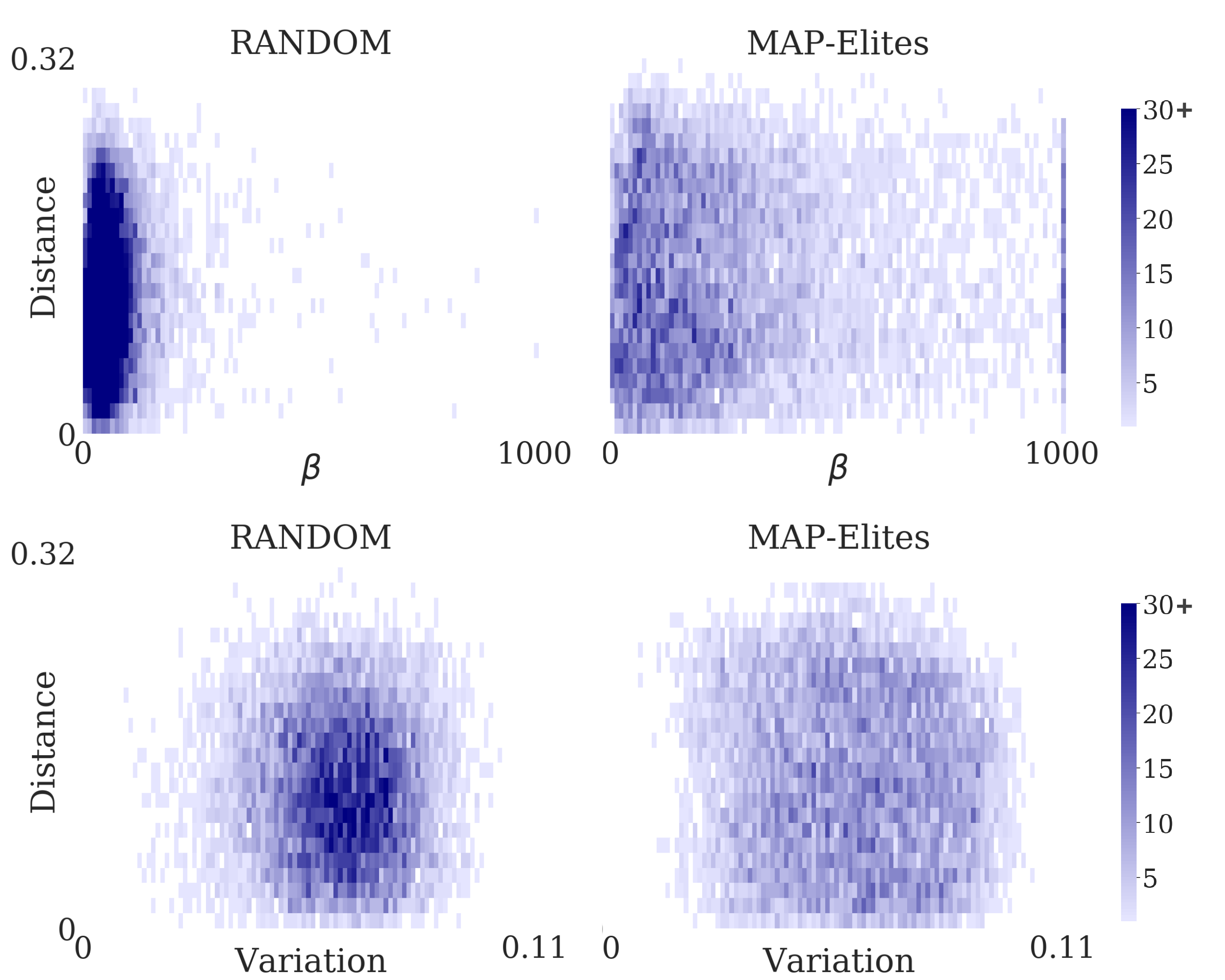}

    \caption{Distribution of cells explored for random search and MAP-Elites. The cell colors represent frequency counts.}
    \label{fig:hist}
    \end{figure}
\subsection{Analysis} \label{subsec:Analysis}
Table~\ref{tab:results} summarizes the performance of the three algorithms, for each of the three behavior spaces. We conducted a two-way ANOVA to examine the effect of the behavior space and the search algorithm on the QD-Score and coverage. There was a statistically significant interaction between the search algorithm and the behavior space for both QD-Score ($F(4,36) = 62.39, p < 0.001$) and coverage ($F(4,36) = 77.92, p < 0.001$). Simple main effects analysis with Bonferroni correction showed that \mbox{MAP-Elites} outperformed \mbox{CMA-ES} and random search in both QD-Score and coverage ($p<0.001$ in all comparisons). This result supports our hypothesis. 

Fig.~\ref{fig:maps} shows the improvement in the QD-Score over time and one example archive from each algorithm for the  first two behavior spaces. MAP-Elites visibly finds more cells and of higher quality (red color), illustrating its ability to cover larger areas of the archive with high-quality scenarios. As expected, CMA-ES concentrates in regions of high-quality scenarios but has small coverage.

Random search covers a smaller area of the archive, compared to MAP-Elites, because of the \textit{behavior space distortion}, shown in Fig.~\ref{fig:hist}. Even though the search parameters are sampled uniformly, scenarios are concentrated on the left side of the archive specified by the human rationality and distance between goals BCs (Fig.~\ref{fig:hist}-top). This occurs because if any of the sampled waypoints deviates from the optimal path, low values of rationality become more likely. In the human variation and distance between goals BCs, the distribution of scenarios generated by random search is concentrated in a small region near the center  (Fig.~\ref{fig:hist}-bottom). This is expected, since the two BCs are Euclidean norms of random vectors (see~\cite{random2018}). On the other hand, MAP-Elites selects elite scenarios from the archive and perturbs them with Gaussian noise, instead of uniformly sampling the scenario parameters, resulting in larger coverage. 




\subsection{Interpreting the Archives}
In the generated archives (Fig.~\ref{fig:maps}), each cell contains an elite, which is the scenario that achieved the maximum assessment value (time to complete the task) for that cell. It is important to confirm that \emph{the timeouts (red cells in the archive) occur because of the implementation of the tested algorithm (hindsight optimization), rather than being artifacts of the simulation environment}.

Therefore, we replay the elites in different regions of the archives to explain the system's performance. We focus on the first two behavior spaces in Fig.~\ref{fig:maps} using the archives generated with MAP-Elites, since MAP-Elites had the largest QD-Score and coverage.




We observe that if the distance between goals is large and the human is nearly optimal, the robot performs the task efficiently. This is shown by the blue color in the top-right of the first behavior space (distance and human rationality $\beta$). We observe the same for large distance and small variation in the second behavior space.

We then explore different types of scenarios where the robot fails to reach the user's goal by the maximum time (10s), indicated by the red cells in the archives. When human variation is large (or equivalently rationality is low), the human may provide inputs that guide the robot towards the wrong goal. Since the robot updates a probability distribution over goals based on the user's input~\cite{javdani2015hindsight} and the robot assumes that the user minimizes their distance to their desired goal, noisy inputs may result in assigning a higher probability to the wrong goal and the robot will move towards that goal instead. Fig.~\ref{fig:elites}(left) shows the execution trace of one elite where this occurs. Fig.~\ref{fig:best} shows the position of this elite in the archive. Fig.~\ref{fig:elites}(center) shows how the robot would reach the desired goal if the human had behaved optimally, instead. 






What is surprising, however, is that the robot does not reach the user's goal even in parts of the behavior space where human variation is nearly 0 (or equivalently rationality is very high), that is when the human provides a nearly optimal input trajectory! Fig.~\ref{fig:elites}(right) reveals a case, where the two goal objects are aligned one closely behind the other. The robot approaches the first object, on the way towards the second object, and stops there.

What is interesting in both scenarios is that the robot gets ``stuck'' at the wrong goal, even when the simulated user continues providing inputs to their desired goal! Inspection of the publicly available implementation~\cite{ada_code} of the algorithm shows that this results from the combination of two factors: the robot's cost function and the human inputs.




\noindent\textbf{Cost Function.} The cost function that the robot minimizes is specified as a constant cost when the robot is far away from the goal and as a smaller linear cost when the robot is near the target~\cite{javdani2018shared} (distance to target is smaller than a threshold). This makes the cost of the goal object near the robot significantly lower than the cost of the other goal objects, which results in the probability mass of the goal prediction to concentrate on that goal. While this can help the user align the end-effector with the object (see~\cite{javdani2018shared}), it can also lead to incorrect inference, if the robot approaches the wrong goal on its way towards the correct goal or because of noisy human input. We confirmed that removing the linear term from the cost function results in the robot reaching the right goal in both examples.



\noindent\textbf{Human Inputs.} The hindsight optimization implementation minimizes a cost function specified as the sum of two quadratic terms, the expected cost-to-go to assist for a distribution over
goals, and a term penalizing disagreement with the user's input. The first-order approximation of the value function leads to an interpretation of the final robot action $u_R = u_R^A+u_R^u$ as the sum of two independent velocity commands, an ``autonomous'' action $u_R^A$ towards the distribution over goals and an action that follows the user input $u_R^u$, as if the user was directly teleoperating the robot (see~\cite{javdani2018shared}).

We have simulated the human inputs, so that they provide a translational velocity command towards the next waypoint, proportional to the distance of the robot's end-effector to that waypoint (section~\ref{subsec:parameters}). This results in a term $u_R^u$ of small magnitude when the end-effector is close to one of the waypoints. If at the same time the robot has high confidence on one of the goals,  $u_R^A$ will point in the direction of that goal and it will cancel out any term $u_R^u$ that attempts to move the robot in the opposite direction.

We confirmed that, if the user instead applied the maximum possible input towards their desired goal, the robot would get ``unstuck,'' so a real user would always be able to eventually reach their desired goal. However, this requires effort from the user who would need to ``fight'' the robot. Overall, the archive reveals limitations that depend on \textit{how the goal objects are aligned in the environment, the direction and magnitude of user inputs, and  the cost function used by the implementation of the hindsight optimization algorithm.}

      \begin{figure}[t!]
        \begin{tabular}{cc}
            \centering
    \begin{subfigure}[t]{.49\columnwidth}
        \includegraphics[width=1.0\columnwidth]{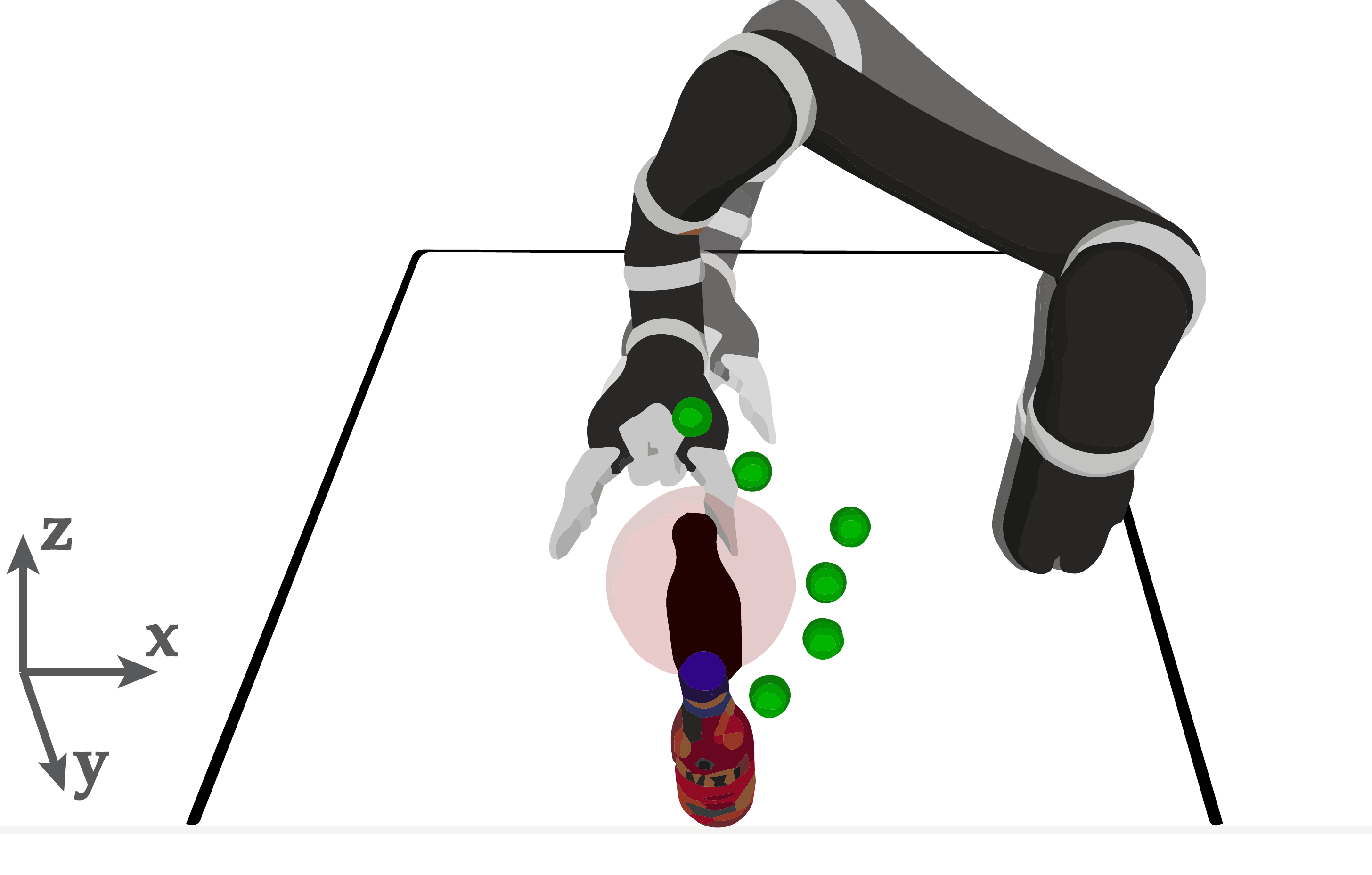}
        \end{subfigure} &
        \begin{subfigure}[t]{0.49\columnwidth}
        \includegraphics[width=1.0\columnwidth]{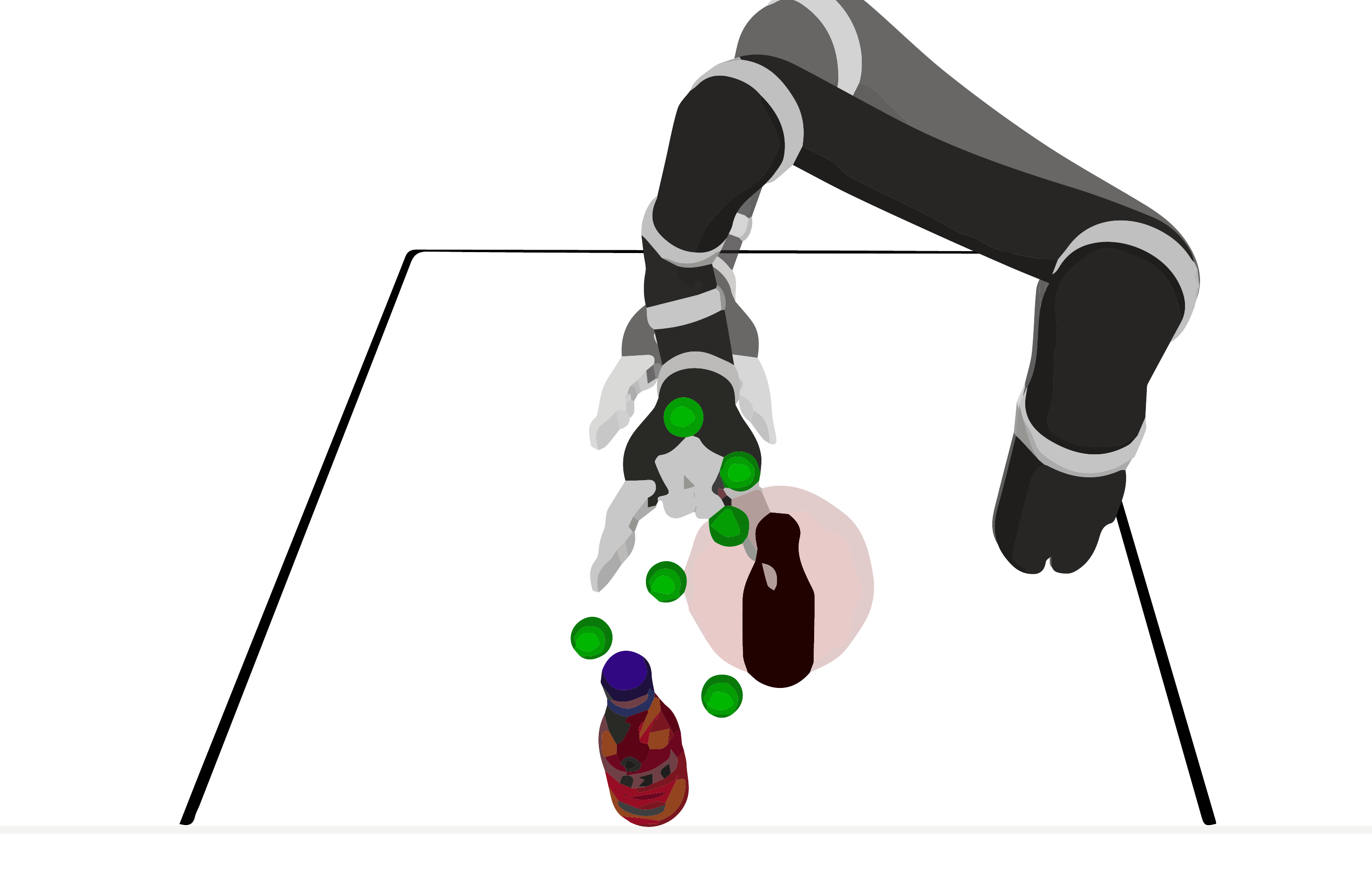}
        \end{subfigure}
        \end{tabular}
                \caption{Scenarios where the policy blending algorithm results in collision with an obstacle, approximated by a sphere. (Left) While the human and robot trajectories are each collision-free, blending the two results to collision when they point towards opposite sides of the obstacle. (Right) Blending with a very noisy human input results in collision. }
    \label{fig:obstacle-examples}
    \end{figure}
    

   \begin{figure}[t!]
    \centering
    \includegraphics[width=0.6\columnwidth]{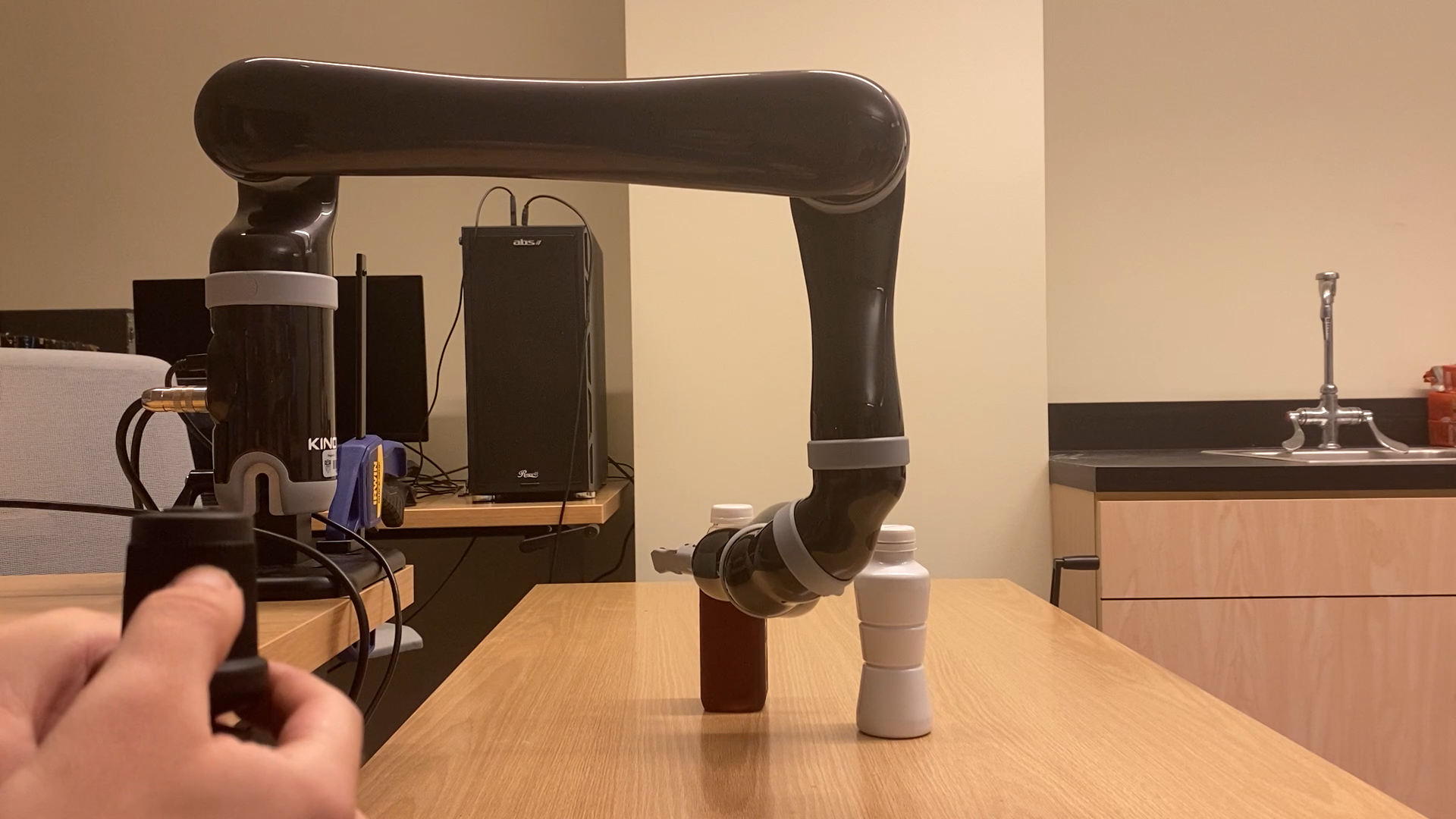}
    \caption{We reproduced the generated scenarios in the real world with actual joystick inputs.}
    \label{fig:real}
    \end{figure}
\section{Comparing Algorithms}\label{sec:comparing_short}
Given the effectiveness of quality diversity in automatically generating a diverse range of test scenarios, we can also use it to understand differences in performance between algorithms. In this work, we compare the performance of  hindsight optimization~\cite{javdani2015hindsight} with linear policy blending~\cite{dragan2012formalizing}. We describe the experiment in Appendix~\ref{sec:comparing}. We found that policy blending resulted in collisions, even for a nearly optimal human, in cases where human and robot inputs pointed towards opposite sides of an obstacle (Fig.~\ref{fig:obstacle-examples}). This confirms previous theoretical findings~\cite{trautman2015assistive} of unsafe behavior that linear blending has in the presence of obstacles. On the other hand, hindsight optimization avoids collisions, since it uses the human  inputs  as  observations  and  the  robot’s  motion is determined only by the robot’s policy.  

\section{Discussion} \label{sec:discussion}

\noindent\textbf{Experimental Findings.} We found that failure scenarios for hindsight optimization occur when the two goals are close to each other and the human inputs are noisy, or when one goal is in front of the other. In the latter case, failure occurs even if the human input is nearly optimal in minimizing the distance to the desired goal. In both cases, the robot becomes over-confident about the wrong goal and gets ``stuck'' there. 

An important factor is the linear decrease of the cost in the vicinity of the goal objects. When specifying the cost function, it would be prudent to make the distance threshold for the linear decrease proportional to the distance between the goal objects, rather than setting it to an absolute value. 

Other potential measures to avoid the system's overconfidence towards the wrong goal are: (1) including the Shannon entropy with respect to all the goals in the cost function~\cite{jeon2020shared} to penalize actions that result in very high confidence to one specific goal; (2) assigning a non-zero probability that the user changes their mind throughout the task and switches goals~\cite{nikolaidis2017human, jain2018recursive}. It would be interesting to investigate the effect of more ``conservative'' assistance on subjective and objective metrics of the robot's performance.

Finally, while linear policy blending naturally gives more control to the user and it is preferred by users in simple tasks~\cite{javdani2018shared}, we empirically verified that the algorithm can generate unsafe trajectories, even if the individual human and robot inputs are safe. 

To show that the presented scenarios can occur in deployed systems, we reproduce them in the real world with actual inputs through a joystick interface (Fig.~\ref{fig:real}).\footnote{We show different generated scenarios reproduced in the real world here: \url{https://youtu.be/2-JCO3dUHsA}} 

\noindent\textbf{Stochasticity in Scenarios.} In our experiments the generated scenarios are deterministic. One may wish to simulate scenarios where there is stochasticity in the robot's decision making or in the environment dynamics. A designer may also wish to test the system's performance under a stochastic human model, e.g., when human inputs are generated by a stochastic noisily rational human.

The most common approach in evolutionary optimization of noisy domains is \textit{explicit averaging}, where we run multiple trials of the same scenario and then retain an aggregate measure of the assessment estimate, e.g, we compute the average to estimate the expected assessment $\mathbb{E}[f(\theta,\phi)]$~\cite{rakshit2017noisy,jin2005evolutionary}. We can follow the same process to estimate the behavior characteristics~\cite{justesen2019map}. To improve the efficiency of the estimation, previous work has also employed implicit averaging methods, e.g., where the assessment of a scenario ($\theta, \phi$) is estimated by taking the assessments of previously evaluated scenarios in the neighborhood of $\theta,\phi$ into account. Previous work also includes adaptive sampling techniques, where the number of trials increases over time as the quality of the solutions in the archive improves~\cite{justesen2019map}. A recent variant of MAP-Elites (Deep-Grid MAP-Elites) which updates a subpopulation of solutions for each cell in the behavior space has shown significant benefits in sample efficiency ~\cite{flageat2020fast}. We leave these exciting directions for future work.

\noindent\textbf{Limitations.} An important challenge is how to effectively characterize  the behavior spaces. While we have assumed bounded behavior spaces, the rationality coefficient does not meet this assumption, which resulted in elites accumulating in the upper bound of the rationality in the archive (Fig.~\ref{fig:maps}). Adapting the boundaries of the space dynamically based on the distribution of generated scenarios~\cite{fontaine:gecco19} could improve coverage in this case.

While distance between objects indeed played a role, our experiments showed the unexpected and surprising edge case where the two objects are in column formation and the human is nearly optimal. An interesting follow-up experiment would be to specify as a BC some metric of object alignment in column formation and investigate further the effect of this variable. In general, a practitioner can test the system with an initial design of BCs, observe the failure cases, create new BCs from newly observed insights and test the system further.
 
 We focused on how to effectively search the generative space of scenarios, but not on the generation methods themselves. Realism is an important future consideration, both in generating environments and human inputs. In human training, realism can be measured through a modified Turing test designed to require humans to distinguish generated scenarios from human authored ones~\citep{martin:ucf}. Alternatively, we could run a user study where we place objects in the same locations as our failure scenarios and observe whether participants perform similar actions that cause failures.

\noindent\textbf{Implications.}
Finding failure scenarios of HRI algorithms will play a critical role in the adoption of these systems. We proposed quality diversity as an approach for automatically generating scenarios that assess the performance of HRI algorithms in the shared autonomy domain and we illustrated a path for future work. While real-world studies are essential in evaluating complex HRI systems, automatic scenario generation can facilitate understanding and tuning of the tested algorithms, as well as provide insights on the experimental design of real-world studies, whose findings can in turn inform the designer for testing the system further. We are excited about applications of quality diversity algorithms as test oracles in verification systems~\cite{kress2020formalizing,porfirio2018authoring}, as well as in other domains where deployed robotic systems face a diverse range of interaction scenarios. 

\section{Acknowledgements}
We thank Tapomayukh Bhattacharjee, David Hsu, Shen Li, Dylan Losey, Dorsa Sadigh, Rosario Scalice, Julian Togelius and our anonymous RSS reviewers for their feedback on early versions of this work.

\bibliographystyle{ACM-Reference-Format}
\bibliography{MarioGANreferences.bib,NSFReferences.bib,references.bib}

\appendices

   \begin{figure*}[t!]
    \centering
    \begin{tabular}{cc}
            \begin{subfigure}[t]{.49\textwidth}
        \centering
        \includegraphics[width=1.0\textwidth]{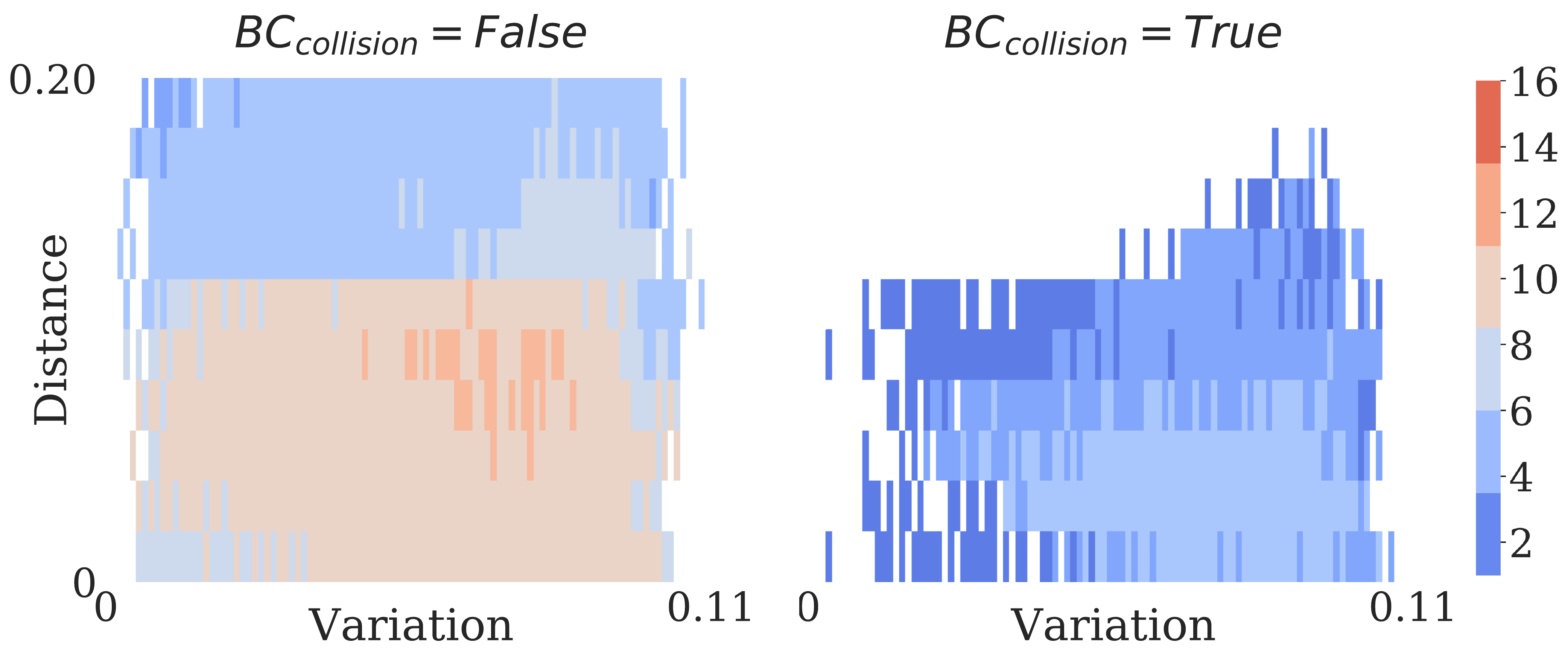}
    \caption{Policy Blending}
        \label{fig:blending}
    \end{subfigure} &
    \begin{subfigure}[t]{.49\textwidth}
        \centering
        \includegraphics[width=1.0\textwidth]{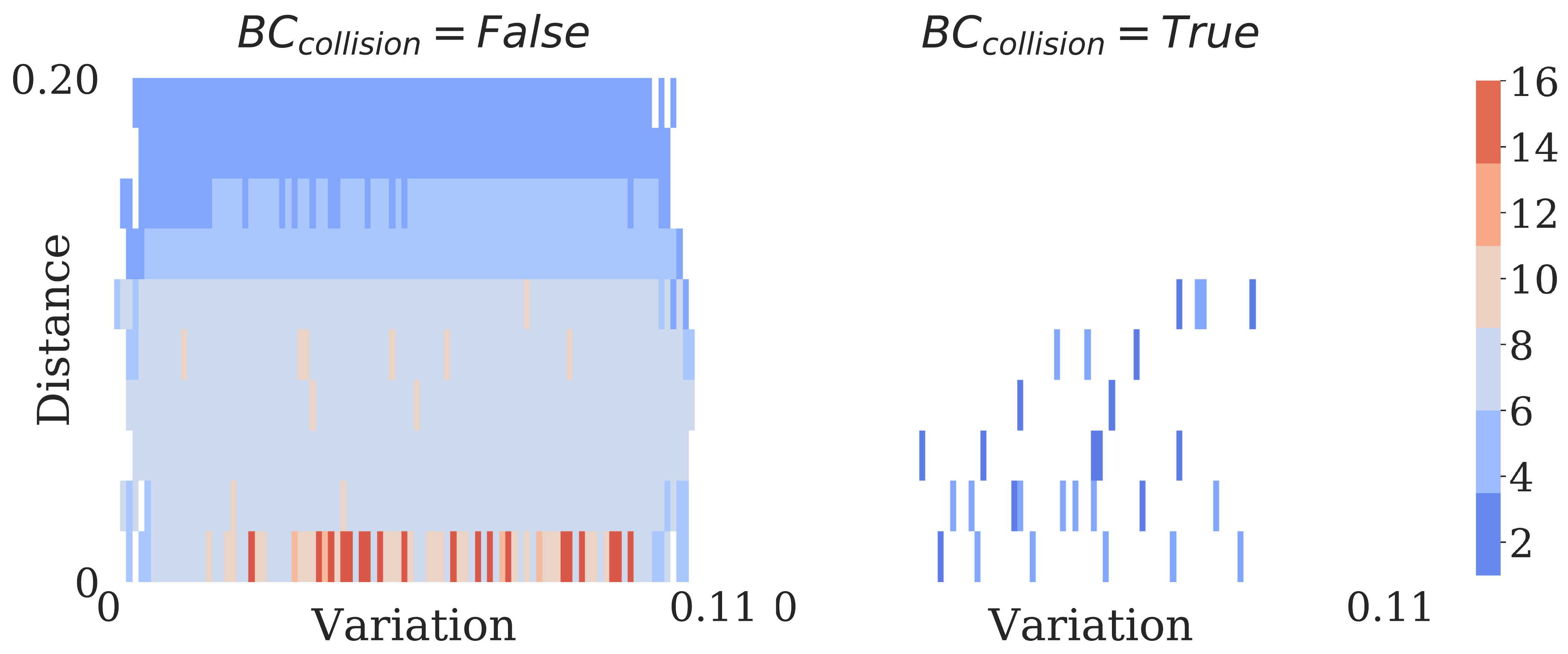}
    \caption{Hindsight Optimization}
        \label{fig:hindsight}
    \end{subfigure}
    \end{tabular}
    \caption{Archives generated with MAP-Elites for the policy blending and hindsight optimization algorithms. For each algorithm, we present the 3D behavior space as two 2D spaces, one where there was no collision ($BC_{collision}=False$), and one where a collision occurred ($BC_{collision}=True$). The colors of the cells in the archives represent duration of the task in seconds.}
    \label{fig:obstacle-map}
    \end{figure*}

\section{Comparing Hindsight Optimization \\ with Policy Blending} \label{sec:comparing}
Given the effectiveness of quality diversity in automatically generating a diverse range of test scenarios, we can also use it to understand differences in performance between algorithms. Using the shared autonomy domain as example, we compare hindsight optimization~\cite{javdani2015hindsight} with policy blending~\cite{dragan2012formalizing}.

\noindent\textbf{Policy Blending.} In shared autonomy, policy blending is a widely used alternative to POMDP-based methods in shared autonomy, since it provides an intuitive interface for combining robotic assistance with user inputs~\cite{carlson2012collaborative,li2011dynamic,gopinath2016human,Muelling2017,dragan2012formalizing,jain2019probabilistic}. In this approach, the robot's and user's actions are treated as two independent sources. The final robot motion is determined by an arbitration function that regulates the two inputs, based on the robot's confidence about the user's goal. A special case of policy blending is \textit{linear blending}. 
Theoretical analysis of linear blending~\cite{trautman2015assistive} has shown that it can lead to unsafe behavior in the presence of obstacles.

We want to empirically confirm these theoretical findings and assess how hindsight optimization and linear policy blending perform in the presence of obstacles. We thus created a new environment with one goal object and a second object that acts as an obstacle. Since there is only one unique goal object, there will be no failure cases arising from incorrect inference as in section~\ref{sec:limiting}.

 In linear blending, robot and human inputs are weighted by the arbitration coefficients $\alpha$ and $1-\alpha$ respectively.  Previous work~\cite{dragan2012formalizing} considers two types: an ``aggressive'' blending, where the robot takes over ($\alpha = 1.0$) if confidence is above a threshold ($conf \geq 0.4$), and a ``timid'' blending where $\alpha$ linearly increases from 0 to 0.6 for $0.4 \leq conf\leq 0.8$. When we have only one goal, confidence is always 1.0 and ``aggressive'' blending and hindsight optimization are identical. Therefore, we consider only the ``timid'' mode which sets $\alpha$ to 0.6.

\noindent\textbf{Obstacle Modeling.} While the hindsight optimization implementation~\cite{ada_code} models the user as minimizing the Euclidean distance between the robot's end-effector and their goal, this is no longer applicable in the presence of obstacles, since a direct path to the goal may collide with the obstacle. Exact computation of the value function is also infeasible, given that the state and action spaces are continuous and the robot acts in real-time. We simplify the computation by modeling the obstacle as a sphere and, if a direct path to the goal intersects with the sphere, we find the intersecting points and approximate the value function as the length of the shortest path to the goal that wraps around the sphere. We do the same for the human waypoints (before adding any disturbances), so that an optimal human will follow the shortest collision-free path to the goal. For robustness, if the end-effector touches the surface of the sphere, it receives an additional velocity command in the direction away from the center of the sphere, similarly to a potential field~\cite{crandall2002characterizing}. We found the sphere approximation simplification adequate for detecting collisions with the hindsight optimization and policy blending algorithms.

\noindent\textbf{Experiment.} We fix the position of the goal object and the obstacle in the y-axis, so that the obstacle $o$ is always between the robot and the goal, and we search for the coordinates ($\phi = (g_x, o_x)$) in the x-axis (Fig.~\ref{fig:obstacle-examples}). The goal object has three target grasp poses. Identically to section~\ref{sec:limiting}, we search for the disturbances to the human inputs ($\theta = (d_1, ... , d_5)$). We specify three BCs: the distance between the two objects in the x-axis, the human variation -- identically to section~\ref{sec:limiting} --  and a binary BC, $BC_{collision}\in \{True,False\}$ indicating whether the robot's end-effector has collided with the obstacle. As before, we set the assessment function $f$ to the time to completion, with a maximum limit of 15$s$. The task terminates if the robot reaches the goal or if it collides with the obstacle. We generate 20,000 scenarios with one run of MAP-Elites to test the two algorithms.

\noindent\textbf{Analysis.} 
Fig.~\ref{fig:obstacle-map} shows the archives generated for the policy blending and the hindsight optimization algorithms. For each algorithm, we present two 2D archives, one with $BC_{collision}=True$ and one with  $BC_{collision}=False$. We compare the archives of the two algorithms in terms of coverage and quality of the scenarios.

\textit{Coverage.} The coverage was 47\% for hindsight optimization, compared to 69\% for policy blending. We observe that the archive with $BC_{collision} = True$ is heavily populated in policy blending (Fig.~\ref{fig:blending}). The archive shows collisions even for a nearly optimal human; while both the human and robot inputs were collision free, sometimes they pointed towards opposite sides of the obstacle, and blending the two resulted in collision (Fig.~\ref{fig:obstacle-examples}(left)). This result matches the theoretical predictions~\cite{trautman2015assistive} of unsafe trajectories in linear blending. 

The $BC_{collision}=True$ archive in Fig.~\ref{fig:blending} also shows that when the horizontal distance exceeds a threshold, the robot can reach the goal with a nearly straight path, and collision occurs only when the human inputs are very noisy (Fig.~\ref{fig:obstacle-examples}(right)).

On the other hand, hindsight optimization resulted in a nearly empty archive when $BC_{collision} = True$. In other words, there were very few scenarios where a collision has occurred. The reason is that hindsight optimization uses the human inputs as observations, and the robot's motion is determined only by the robot's policy; the few sparse collisions are artifacts of the OpenRAVE environment, occurring because of infrequent lags in sending velocity commands to the robot.

\textit{Scenario Assessment.} When $BC_{collision} = False$, the average time to completion was 7.95s for policy blending and 6.70s for hindsight optimization, indicating that policy blending took more time to finish the task. This result matches previous human subjects experiments~\cite{javdani2018shared}, where policy blending took longer than hindsight optimization, and it was caused by two factors: (1) Often the human and robot inputs would point to opposite directions, resulting in velocities of small magnitudes when blended. (2) The human inputs were proportional to the distance to the next waypoint, spiking at the waypoints and decreasing until the next waypoint (section~\ref{subsec:parameters}). This resulted in parts of the trajectory where the human inputs were of smaller magnitude than the robot's, so blending the two gave smaller velocity commands than treating the human inputs as observations. The red bars in hindsight optimization (Fig.~\ref{fig:hindsight}) indicate timeouts; when the goal object was exactly behind the obstacle, the robot sometimes oscillated between taking a grasping position and moving away from the sphere. 

Overall, the generated scenarios show two potential limitations of linear policy blending, compared to hindsight optimization: unsafe behavior in the presence of obstacles, and delays in task completion, when human and robot inputs point to opposite directions.

\end{document}